\documentclass[final]{cvpr}

\usepackage{times}
\usepackage{epsfig}
\usepackage{graphicx}
\usepackage{amsmath}
\usepackage{amssymb}
\usepackage{multirow}
\usepackage{algorithm}
\usepackage{algpseudocode}
\usepackage{amsmath}
\usepackage{makecell}
\usepackage[pagebackref=false,breaklinks=true,colorlinks,bookmarks=false]{hyperref}

\def\cvprPaperID{} % *** Enter the CVPR Paper ID here
\def\confYear{CVPR 2021}

\begin{document}

%%%%%%%%% TITLE
\title{Action Unit Memory Network for Weakly Supervised Temporal Action Localization }

\author{Wang Luo$^1$, Tianzhu Zhang\textsuperscript{1,\thanks{Corresponding Author}}, Wenfei Yang$^1$, Jingen Liu$^2$, Tao Mei$^2$, Feng Wu$^1$, Yongdong Zhang$^1$\\
	%$^1$ School of Information Science and Technology, University of Science and Technology of China\\
	%$^2$ Kuaishou Technology\\
	$^1$ University of Science and Technology of China; $^2$ JD AI Research\\
	{\tt\small \{lw1998,yangwf\}@mail.ustc.edu.cn; \{tzzhang,fengwu,zhyd73\}@ustc.edu.cn;}\\ {\tt\small jingenliu@gmail.com; tmei@live.com}
	% For a paper whose authors are all at the same institution,
	% omit the following lines up until the closing ``}''.
	% Additional authors and addresses can be added with ``\and'',
	% just like the second author.
	% To save space, use either the email address or home page, not both
}

\maketitle
\pagestyle{empty}  % no page number for the second and the later pages
\thispagestyle{empty} % no page number for the first page

%%%%%%%%% ABSTRACT
\begin{abstract}

Weakly supervised temporal action localization aims to
detect and localize actions in untrimmed videos
with only video-level labels during training.
However, without frame-level annotations, 
it is challenging to achieve localization completeness
%complete localization 
and relieve background interference.
In this paper, we present an Action Unit Memory Network (AUMN) for
weakly supervised temporal action localization,
which can mitigate the above two challenges by learning an action unit memory bank.
In the proposed AUMN,
two attention modules are designed to update the memory bank adaptively
and learn action units specific classifiers.
Furthermore, three effective mechanisms 
(diversity, homogeneity and sparsity) are 
designed to 
guide the updating of
the memory network.
To the best of our knowledge, this is the first work to explicitly model the action units with a memory network.
Extensive experimental results on two standard benchmarks
(THUMOS14 and ActivityNet) demonstrate that our AUMN
performs favorably against state-of-the-art methods. 
Specifically, the average mAP of IoU
thresholds from 0.1 to 0.5 on the THUMOS14 dataset is significantly
improved from 47.0\% to 52.1\%.

\end{abstract}

%%%%%%%%% BODY TEXT
\vspace{-5mm}
\section{Introduction}
\vspace{-7pt}
\label{sec:introduction}
Temporal action localization (TAL) is an important yet challenging task for video understanding. Its goal is to localize temporal boundaries of actions with specific categories in untrimmed videos~\cite{cviu2017thumos,tpami2013temporal}.
%}
%
Because of its broad applications in high-level tasks such as video surveillance~\cite{cvpr2017spatiotemporal},
video summarization~\cite{cvpr2012discovering}, and
event detection~\cite{Kwak_2013_CVPR},
TAL has recently drawn increasing attentions from the community.
Up to now, deep learning based methods have made impressive progresses in this area.
However, most of them handle this task in a fully supervised way,
requiring massive temporal boundary annotations for actions~\cite{cvpr2019gaussian,iccv2019graph,cvpr2018tal-net,iccv2017r-c3d,cvpr2016msc}.
%cvpr2016end,
Such manual annotations are expensive to obtain,
which limits  the development potential of fully-supervised methods in real-world scenarios.
\begin{figure}[t]
\centering
% Requires \usepackage{graphicx}
\includegraphics[width=0.98\linewidth]{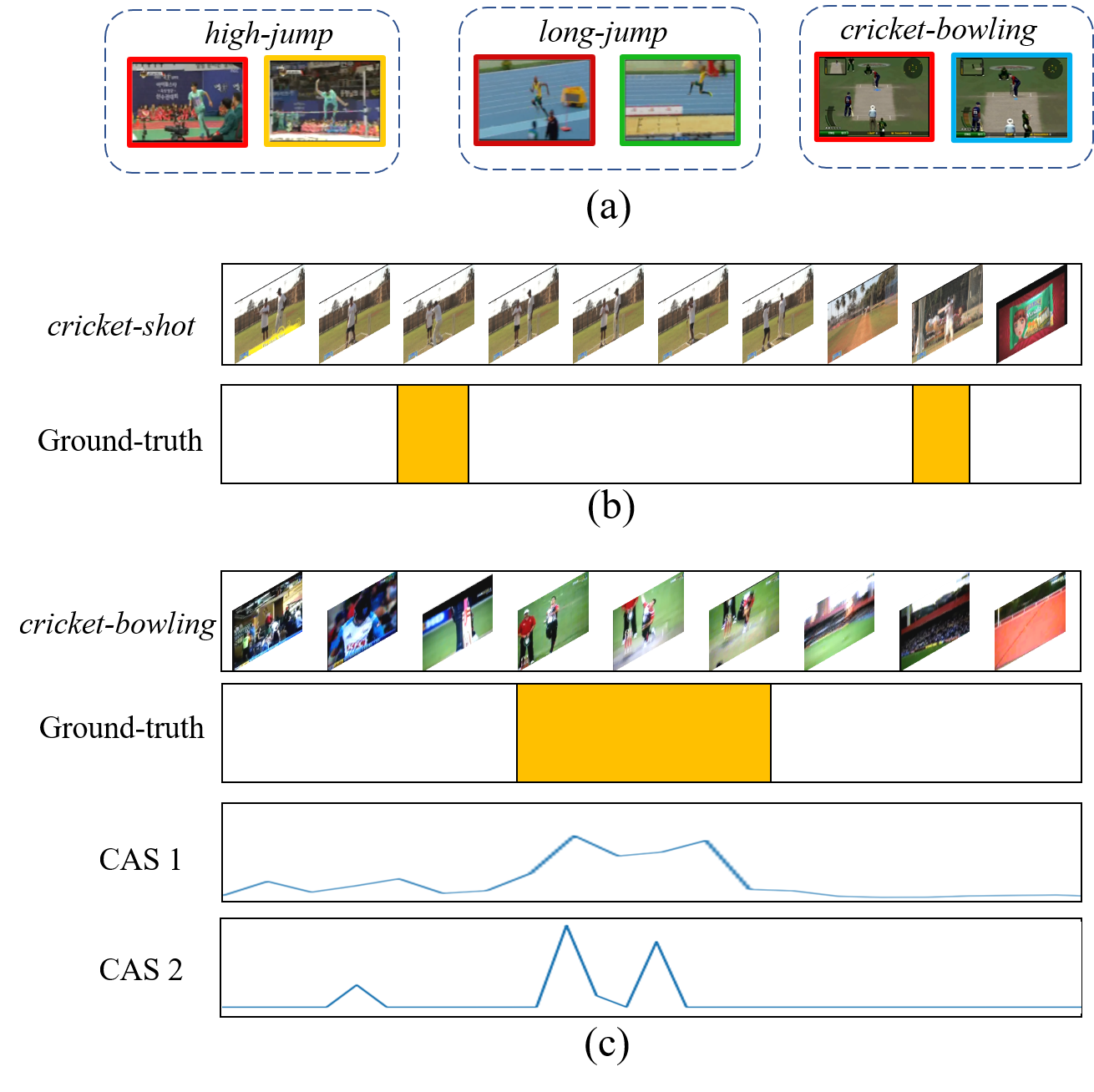}
\caption{
	(a) Illustration of ``Sharing Units'' characteristic. The running (red box) is a shared action unit among \emph{high-jump}, \emph{long-jump} and \emph{cricket-bowling}.
	(b) Illustration of ``Sparsity'' characteristic. An action usually occupies a small portion of untrimmed videos.
	(c) Illustration of ``Smoothness'' characteristic. CAS1 is more suitable for the action localization task because the CAS2 tends to divide a continuous action into multiple instances.
}
\label{fig:motivation}
\vspace{-17pt}
\end{figure}

To relieve this problem, the weakly supervised setting that only requires video-level category labels is proposed~\cite{iccv2017hide-and-seek,MM2018step-by-step-erase,cvpr2017untrimmednets,iccv2017hide-and-seek,MM2018step-by-step-erase,MM2019assg,cvpr2018stpn,iccv2019backgroundmodel,yang2021cvpr, yang2021local}.
It can be formulated as a multiple instance learning problem,
where a video is treated as a bag of multiple segments and fed into a video-level classifier to get a class activation sequence (CAS).
There are two primary challenges,
named localization completeness and background interference.
To solve the first challenge,
previous works usually adopt a well-designed erasing strategy~\cite{iccv2017hide-and-seek,MM2018step-by-step-erase,MM2019assg} or a multi-branch architecture~\cite{cvpr2019cmcs}.
Both of them aim to force the model to concentrate on different parts of videos and hence discover the whole action without missing any relevant segments.
To handle the second challenge, some methods~\cite{eccv2018wtalc,iccv20193cNet,aaai2020relational} employ an attention-based per-class feature aggregation scheme,
where the class-specific attention is obtained by normalizing the CAS along the temporal axis.
This scheme helps learn a compact intra-class feature, which enables action segments to be more discriminative than the background.
Furthermore, one popular way to handle both challenges is to learn a class-agnostic attention mechanism~\cite{cvpr2018stpn,iccv2019backgroundmodel,aaai2020background,cvpr2020weakly,cvpr2020TCAM}, to highlight action segments and suppress background segments.
%}

\vspace{-1mm}
By studying all previous TAL methods, we sum up the following three important observations (\emph{i.e.}, TAL Properties):
(1) Sharing Units. An action to be detected generally consists of some primary action units, which can be shared with other action classes.
For example, as shown in Figure~\ref{fig:motivation} (a), a ~\emph{high-jump} contains running and jumping upward while a ~\emph{long-jump} consists of running and jumping forward,
so running is a shared action unit.
(2) Sparsity. In general, only a sparse set of video segments contains the meaningful target actions.
As we can see from Figure~\ref{fig:motivation} (b),
%As shown in Figure~\ref{fig:motivation} (b),
an action only occupies a small portion of the video.
(3) Smoothness. A smooth CAS is required for localization, because an action is continuous, as shown in Figure~\ref{fig:motivation} (c).
These three characteristics are critical for the success of action localization. Unfortunately, they have not been thoroughly addressed by previous studies.
%}
%{\color{blue}
To achieve accurate and complete action localization,
these three observations should be taken into consideration
%when developing an action localization model.
when designing an action localization model.
However, with only video-level
%category
labels, it is difficult to model them jointly in a unified model.
%}

%{\color {blue}
To fully leverage the above three characteristics for action localization,
we propose a novel end-to-end framework, called Action Unit Memory Network (AUMN), for more effective weakly supervised action localization.
Our framework starts with the action unit templates learning.
According to the ``Sharing Units'' characteristic, we design a sub-network as a memory bank of action unit templates, which serve as our learning primary for action localization.
To exploit the templates for action classification, we further design a Multi-Layer Perceptron (MLP) network to embed each template into the action class space.
Basically, the MLP network helps connect templates to action classes. Intuitively speaking,   action unit templates will be projected onto a set of action classifiers.
Afterwards, a cross-attention module is proposed to compute the relationships between a video segment and all templates.
And according to the ``Smoothness'' characteristic, we introduce a self-attention module to compute the relationships between different segments in a video for aggregating context information.
Leveraging both of the attention mechanisms, we can get refined segment features and be able to dynamically select action classifiers for each video segment,
which in turn, simultaneously contribute the adaptive learning of the memory bank.
%}

However, the video-level ground-truth supervision alone is not enough for memory updating.
Based on the property of action units and  ``Sparsity'' characteristic,
we further design three effective mechanisms to guide the updating of the memory bank:
(1) Since action units are different from each other,
each template in the memory bank should be unique.
To achieve this goal,
we design a diversity mechanism to encourage the differences among the templates in the memory.
(2) While the diversity mechanism can encourage each template in the memory to be unique,
it does not guarantee that no template is useless,
which means that a template may have low similarities with all video segments.
To avoid this, we design a homogeneity mechanism to encourage a uniform distribution for the occurring probability of templates.
(3) In an untrimmed video,
action segments only occupy a small portion of the whole video,
and most of the video segments are background.
Thus we design a sparsity mechanism to encourage that only a sparse set of video segments can have high similarities with the templates in the memory.
These three mechanisms together with the supervision of video-level category label can guide the network to learn meaningful action units.

To sum up, the main contributions of our work are three-fold:
(1) To the best of our knowledge, we are the first to model the action units with a memory network for the weakly supervised TAL task.
(2) We propose two attention modules to ensure our memory to update adaptively and learn action units specific classifiers. Further, three effective mechanisms (diversity, homogeneity and sparsity) are designed to guide the updating.
(3) Extensive experimental results on two challenging benchmarks including THUMOS14~\cite{cviu2017thumos} and ActivityNet~\cite{cvpr2015activitynet} demonstrate that the proposed AUMN performs favorably against state-of-the-art weakly supervised TAL methods.

\vspace{-4mm}
\section{Related Work}
\vspace{-7pt}
In this section,
we overview methods that are related to fully and weakly supervised temporal action localization and memory networks.

\vspace{-1mm}
{\bfseries Fully Supervised Temporal Action Localization}.
Temporal action localization (TAL) aims to not only recognize actions in untrimmed videos but also give an accurate temporal proposal for each action, which makes it very challenging.
To tackle this problem, fully supervised based methods have been extensively studied recently, where the frame-level annotations are required during training~\cite{cvpr2016slidingwindow,cvpr2016msc,cvpr2018tal-net,iccv2017ssn,cvpr2016end,bmvc2017end,iccv2019graph}.
Most of these methods borrow intuitions from the object detection frameworks~\cite{tpami2015rcnn,nips2015faster-rcnn,eccv2016ssd,girshick2014rich,redmon2016you}.
In specific, many methods adopt a two-stage pipeline, \emph{i.e.}, action proposals are generated first and then fed into a classification module.
For proposal generation, some methods adopt the sliding window~\cite{cvpr2016slidingwindow,cvpr2016msc,yang2017exploring,xiong2017pursuit} and others predict temporal boundaries of action instances directly~\cite{cvpr2018tal-net,iccv2017ssn,cvpr2017sst,lin2018bsn,lin2019bmn}.
In addition to the two-stage methods, one-stage methods are proposed to predict action category and temporal boundaries from raw data directly, which are more flexible and efficient~\cite{cvpr2016end,bmvc2017end,cvpr2019gaussian,mm2017ssad}.
\vspace{-1mm}
{\bfseries Weakly Supervised Temporal Action Localization}.
Weakly supervised methods tackle the same problem
but with less supervision, \emph{e.g.}, video-level category labels.
This pipeline can alleviate the requirement
for expensive action boundary annotations,
but raise two challenges named localization
completeness and background interference.
To handle the two problems, existing methods can be divided into three types.
The first type of works attempt to solve
the localization completeness by applying a well-designed erasing strategy~\cite{iccv2017hide-and-seek,MM2018step-by-step-erase,MM2019assg} or a multi-branch architecture~\cite{cvpr2019cmcs}.
For example, Zhong et al.~\cite{MM2018step-by-step-erase} design a step-by-step erasion approach to train the one-by-one classifiers,
via collecting detection results from these classifiers, more action segments are found.
And in CMCS~\cite{cvpr2019cmcs},
a multi-branch network with a diversity loss
is proposed to make the model focus on different parts of videos.
The second type of works aim to
tackle the background interference via a intra-class feature compactness scheme~\cite{eccv2018wtalc,iccv20193cNet,aaai2020relational,eccv2020A2CL-PT}.
They first compute the class-specific attention
by applying the softmax function to CAS
then use this attention to get an aggregated video-level feature.
By devising different mechanisms to
learn a compact intra-class feature,
action and background segments tend to be separated.
For example, to decrease the intra-class variance,
3C-Net~\cite{iccv20193cNet} and A2CL-PT~\cite{eccv2020A2CL-PT}
maintain a set of class center and
RPN~\cite{aaai2020relational} learns class-specific prototypes.
The third type of works are based on a class-agnostic attention mechanism~\cite{cvpr2018stpn,iccv2019backgroundmodel,aaai2020background,cvpr2020weakly,cvpr2020TCAM,eccv2020TSCN,eccv2020EM-MIL}, which can consider both the challenges simultaneously.
Unlike the second type of works,
the attention here is generated in a bottom-up
way from the raw data and trained for
highlighting foreground segments.
It is first proposed by STPN~\cite{cvpr2018stpn}
and then inspires many following methods.
Some of them introduce an auxiliary category to focus on modeling background~\cite{iccv2019backgroundmodel,aaai2020background}.
And based on the observation that background features
differ  from action features,
DGAM~\cite{cvpr2020weakly} adopts
a conditional variation auto-encoder to
construct different feature distributions
conditioned on the attention.
Recently, TSCN~\cite{eccv2020TSCN} and EM-MIL~\cite{eccv2020EM-MIL}
fuse the output of different modalities (RGB and optical flow)
to generate pseudo labels for guidance of the attention.

\vspace{-1mm}
{\bfseries Memory Networks}.
Memory networks typically involve an internal memory
implicitly updated in a recurrent process, e.g., LSTM~\cite{hochreiter1997long}, or an explicit memory bank that can be read or written with an attention based mechanism.
Memory networks that can be trained end-to-end
are first proposed in the natural language processing
research like question answering~\cite{miller2016key} and sentiment analysis~\cite{chen2017recurrent}.
Recently, in the temporal action localization task, a popular use of memory is exploring the temporal structure based on the LSTM~\cite{yeung2018every,cvpr2016end,sun2015temporal}. The ability of LSTM to learn from long sequences with unknown size of background is well-suited for fine-grained action localization from untrimmed videos.
Instead of exploiting the temporal relationships in a video, we propose an attention-based memory mechanism to model the action units which are shared among all the videos. This mechanism helps us to deal with the large intra-class variations, so that we can get more complete localization results by discovering various action units.

\vspace{-4mm}
\section{Our Proposed Approach}
\vspace{-7pt}
\begin{figure*}
\centering
% Requires \usepackage{graphicx}
\includegraphics[width=0.98\linewidth]{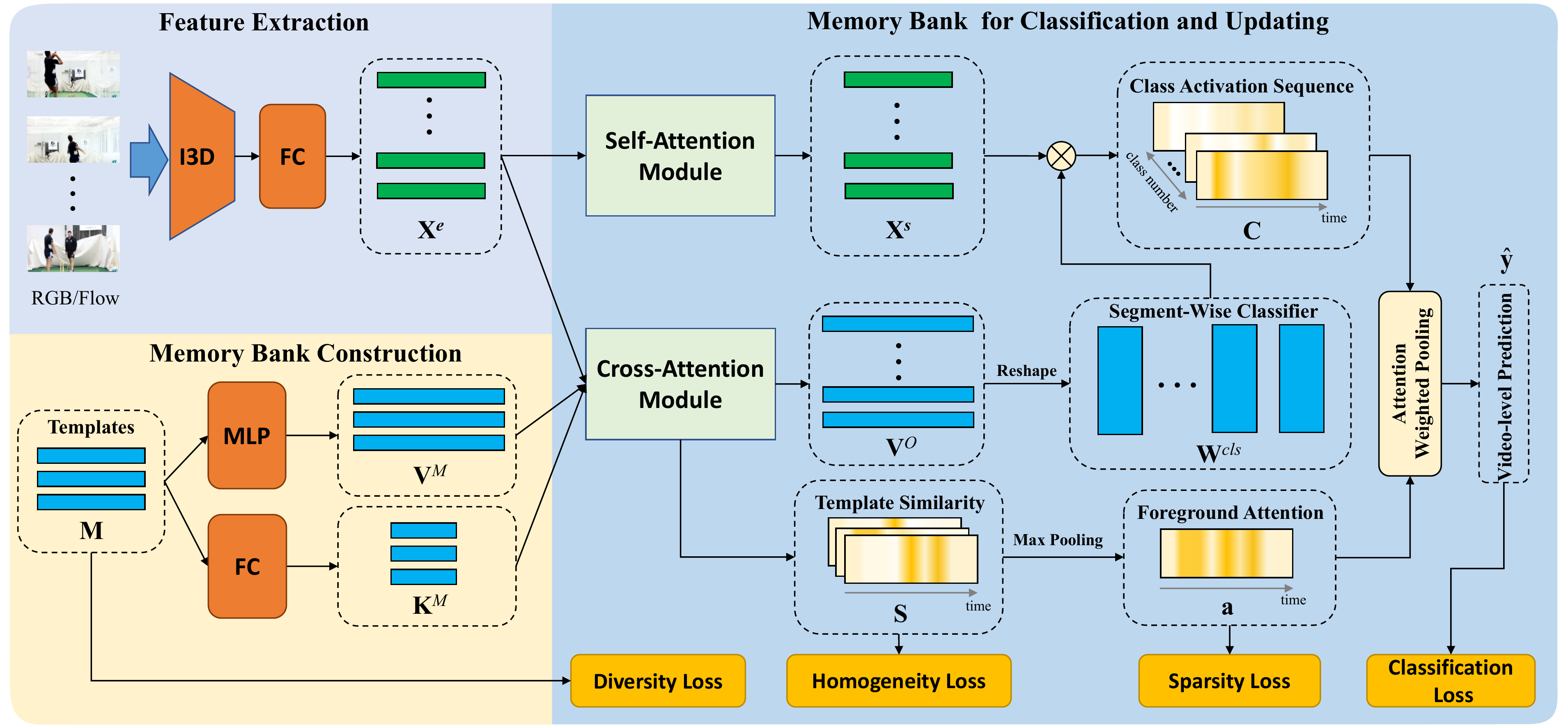}
\caption{
	Overall architecture of our proposed Action Unit Memory Network (AUMN), which consists of three parts: feature extraction, memory bank construction, memory bank for classification and  updating.
}
\label{fig:framework}
\vspace{-6mm}
\end{figure*}
In this section, we first formulate the task of weakly supervised Temporal Action Localization. Then we describe each composition of the proposed Action Unit Memory Network (AUMN) in details.

\vspace{-2mm}
\subsection{Notations and Preliminaries}
\vspace{-2mm}
\label{sec:notation}
Assume we have $N$ untrimmed training videos $\{v_{i}\}_{i=1}^{N}$.
Each video $v_{i}$ has its ground-truth label ${\bf y}_{i} \in \mathbb{R}^{C}\ $,
where $C$ is the number of action categories. ${\bf y}_{i}(j)=1$ if the action category $j$ is present in the video and ${\bf y}_{i}(j)=0$ otherwise\footnote{If there are multiple action categories in one video, ${\bf y}_{i}$ is normalized with the $\ell_1$ normalization.}.
\vspace{-1mm}
During testing, the goal of the temporal action localization is to generate a set of action proposals $\{(c, s, e ,q)\}$ for each video,
where $c$ and $q$ denote the predicted category and the confidence score, $s$ and $e$ represent the start and the end time respectively.
In this paper, we follow previous works~\cite{cvpr2018stpn,iccv20193cNet,cvpr2020weakly} to extract features for both RGB and optical flow streams.
Given an untrimmed video $v_{i}$, we first divide it into non-overlapping 16-frame segments and apply the I3D pretrained on the Kinetics dataset to extract features for each segment.
Then we get two segment-wise features ${\bf X}^{RGB}_i \in \mathbb{R}^{{l_i}\times{D}}$ and ${\bf X}^{FLOW}_i \in \mathbb{R}^{{l_i}\times{D}}$, where $l_i$ denotes the number of segments in video $v_{i}$ and $D$ is the dimension of features.
Because the RGB and FLOW streams are trained independently,
we use ${\bf X}_i$ to represent them in the rest of this paper for simplicity.
Since the extracted features from I3D are learned for the action recognition task originally,
it is desired to add a task-adaption layer to refine the extracted features.
In specific, we adopt a temporal convolutional layer with the ReLU activation as
\begin{equation}
\label{eq:embedding}
\setlength{\abovedisplayskip}{3pt}
\setlength{\belowdisplayskip}{3pt}
  {\bf X}^{e}_i = \mathrm{ReLU}(W^{emb} * {\bf X}_i + b^{emb}),
\end{equation}
where the $*$ represents the convolution operation, $W^{emb}$ and $b^{emb}$ are the weights and bias of temporal filters, ${\bf X}^{e}_i \in \mathbb{R}^{{l_i}\times{F}}$ is the learned embedding feature, and $F$ is the dimension of learned features.

\vspace{-2mm}
\subsection{Action Unit Memory Network}
\vspace{-2mm}
\label{sec:aum}
%As shown in Figure ~\ref{fig:motivation}, each action is a temporal composition of several kinds of action units.
%%
%By modeling action units as different templates that are shared among all the action categories, in the testing stage, we can utilize these templates to find all the action segments in a video and get a complete localization result.
%
The overall architecture of our action unit memory network
is shown in Figure~\ref{fig:framework}.
%
%It consists of three parts including a memory bank, a self-attention module and a cross-attention module.
%
It consists of three parts including
feature extraction, memory bank construction, memory bank for classification and  updating.
The details are introduced as follows.

{\bf Memory Bank Construction}.
The action unit memory bank stores multiple templates ${\bf M} \in \mathbb{R}^{{K}\times{F}}$, and each template represents an action unit,
where $K$ and $F$ are the number and dimension of templates respectively.
We adopt two encoders named ${\bf Enc}_K$ and ${\bf Enc}_V$ to embed the templates into pairs of keys and values respectively.
The ${\bf Enc}_K$ is designed to reduce the dimension of the templates for efficient reading from the memory, which is implemented as a fully connected layer (FC).
And the ${\bf Enc}_V$ is designed to encode each template into a template-specific classifier,
which is a MLP network consisted of two FC layers with a bottleneck structure among them to reduce parameters.
In this way, the keys store appearance and motion related information for the templates and can be used for template matching during memory reading,
and the values store templates specific classifiers and can be used for segment classification.
%}
Formally, we formulate the encodings as follows:
\begin{equation}
\label{eq:enck}
\setlength{\abovedisplayskip}{3pt}
\setlength{\belowdisplayskip}{3pt}
  {\bf K}^{M} = {\bf Enc}_K({\bf M}),
\end{equation}
\begin{equation}
\label{eq:enkv}
  {\bf V}^{M} = {\bf Enc}_V({\bf M}),
\end{equation}
where ${\bf K}^{M} \in \mathbb{R}^{{K}\times{F/m}}$ and ${\bf V}^{M} \in \mathbb{R}^{{K}\times{CF}}$ are keys and values, $M$ denotes the memory, and $m$ is a hyper-parameter to control memory reading efficiency.
%{\color{cyan}
Given the memory bank and an input video, we introduce how to perform video classification and memory updating next.
%}
%
%{\color{red}
%For convenience of reshaping operation of ${\bf V}^{M}$ in the later method, we choose $n$ to be equal to the action category amount $C$.
%}
%{\color{blue}
%delete this sentence
%%And and each value in ${\bf V}^{M}$ can be reshaped into $R^{C\times F}$ for later classification.
%}
%

%
\vspace{-1mm}
{\bf Memory Bank for Classification}.
For video classification,
we use an encoder ${\bf Enc}_Q$ which is implemented as a FC layer to encode video feature ${\bf X}^e_{i}$ into a set of queries ${\bf Q}_i \in \mathbb{R}^{{l_i}\times{F/m}}$,
and then feed the segment features and queries into a self-attention module and a cross attention module to generate classification results.
In the self-attention module, we first calculate the similarity scores among video segments with queries and then use these scores to refine the segment features by aggregating context information, which can be formulated as
%}
\begin{equation}
\label{eq:self attention}
\setlength{\abovedisplayskip}{3pt}
\setlength{\belowdisplayskip}{3pt}
  {\bf X}^s_i = (\mathrm{softmax}(\frac{{\bf Q}_i {\bf Q}^T_i}{\sqrt{F/m}}) + {\bf I}){\bf X}^e_i,
\end{equation}
where ${\bf I}$ is the identity matrix used to preserve the original information, and ${\bf X}^s_i$ keeps the same dimension with ${\bf X}^e_i$. Via this message passing between segments, we can extract global context information and get more discriminative features for both classification and localization.
%

%
%{\color{cyan}
In the cross-attention module,
we read from the memory and get a set of segment-wise classifiers.
To achieve this goal,
we first calculate the similarity scores   ${\bf S}_i \in \mathbb{R}^{{l_i}\times{K}}$ between video segments and memory templates with the scaled dot-product
\begin{equation}
\label{eq:cross attention}
\setlength{\abovedisplayskip}{3pt}
\setlength{\belowdisplayskip}{3pt}
  {\bf S}_i = \mathrm{sigmoid}(\frac{{\bf Q}_i ({\bf K}^M)^T}{\sqrt{F/m}}).
\end{equation}
Based on the similarity scores,
we can obtain a set of segment-wise classifiers by using the
similarity scores to aggregate memory values as
\begin{equation}
\label{eq:message retrieval}
\setlength{\abovedisplayskip}{3pt}
\setlength{\belowdisplayskip}{3pt}
  {\bf V}^O_i = {\bf S}_i{\bf V}^M,
\end{equation}
where ${\bf V}^O_i \in \mathbb{R}^{{l_i}\times{CF}}$.
Later, to perform classification, we reshape ${\bf V}^O_i$ into a set of segment classifiers ${\bf W}^{cls}_i = \{{\bf W}^{cls}_i(t) \in \mathbb{R}^{{F}\times{C}}\}_{t=1}^{l_i}$, which is adaptive to the appearance or motion variations of each segment.

\vspace{-2pt}
With the refined features of the self-attention module and the segment-wise classifiers of the cross-attention module,
we can obtain the segment-level classification results by applying each classifier on the corresponding segment.
%
%{\color {red}
%The segment-level classification results are then aggregated into video-level classification result $\hat{\bf y}_i$ by attention weighted pooling
%}
%%
%{\color {blue}
Since we only have video-level ground-truth supervision,
we need to aggregate these segment-level classification results into a video-level prediction.
%
%{\color {red}
%In specific, the similarity matrix ${\bf S}_i$ is   pooled over $K$ into the class-agnostic weights  first}
%{\color{cyan}
In specific, since the second dimension of the similarity matrix ${\bf S}_i$ denotes the similarity between a segment and a template,
we can apply the max-pooling operation along the second dimension of the ${\bf S}_i$ to get the foreground attention weight ${\bf a}_i \in \mathbb{R}^{l_i}$ as
\begin{equation}
\label{eq:class agnostic weights}
\setlength{\abovedisplayskip}{3pt}
\setlength{\belowdisplayskip}{3pt}
  {\bf a}_i = \mathrm{MaxPool}({\bf S}_i),
\end{equation}
%
%where ${\bf a}_i \in \mathbb{R}^{l_i}$ represents the foreground attention of each segment.delete this }
%
The video-level classification result $\hat{\bf y}_i$ is then obtained as an attention weighted pooling
%}
%
\begin{equation}
\label{eq:prediction}
\setlength{\abovedisplayskip}{3pt}
\setlength{\belowdisplayskip}{3pt}
\hat{\bf y}_i = \mathrm{softmax}(\frac{1}{l_i}\sum_{t=1}^{l_i}{\bf a}_i(t)({\bf X}^s_i(t){\bf W}^{cls}_i(t))),
\end{equation}
where $\hat{\bf y}_i \in \mathbb{R}^{C}$.
Then the classification loss is defined as the cross-entropy loss between the prediction and the video label ${\bf y}_i$
\vspace{-1mm}
\begin{equation}
\label{eq:cls}
\setlength{\abovedisplayskip}{3pt}
\setlength{\belowdisplayskip}{3pt}
\mathcal{L}_{cls} = - \frac{1}{B}\sum_{i=1}^{B}\sum_{j=1}^{C}{\bf y}_i(j) \log \hat{\bf y}_{i}(j).
\end{equation}

\vspace{-1mm}
{\bf Memory Bank Updating.}
%{\color{red}
%However, we find that the supervision of video-level category label is not enough to learn a satisfying memory bank.
%}
%{\color{cyan}
For the memory updating, we find that the above classification loss alone is not enough to learn a satisfying memory bank.
%}
%
Thus we design three mechanisms (diversity, homogeneity and sparsity) to guide the updating of the memory bank.
The diversity mechanism encourages that each template in the memory bank is different from other templates,
the homogeneity mechanism encourages that each template in the memory bank is meaningful,
and the sparsity mechanism encourages that the templates in the memory bank can suppress background segments.
In specific, in the diversity mechanism, we design a diversity loss to ensure the uniqueness of each template in the memory as
\begin{equation}
\label{eq:d_loss}
\setlength{\abovedisplayskip}{3pt}
\setlength{\belowdisplayskip}{3pt}
  \mathcal{L}_{d} = \left\|{\bf M}{\bf M}^T - {\bf I}\right\|_F,
\end{equation}
where ${\bf I}$ is the identity matrix and $\left\|\cdot\right\|_F$
is the Frobenius norm of a matrix.
While the diversity loss encourages the templates in the memory bank to be unique,
it does not guarantee that each template in the memory bank is useful.
For example, a template may not represent an action unit and have low similarities with all video segments during training.
To deal with this issue,
we design a homogeneity loss to encourage a
uniform distribution for the occurring probability of
templates in the homogeneity mechanism.
In specific,
we first pool the similarity matrix ${\bf S}_i$ over time
by a sum operation and then use a softmax function to
obtain the occurring probability of each template as
\begin{equation}
\label{eq:occurrence probability}
\setlength{\abovedisplayskip}{3pt}
\setlength{\belowdisplayskip}{3pt}
  {\bf p}^O_i = \mathrm{softmax}(\sum_{t=1}^{l_i}{\bf S}_i(t)),
\end{equation}
where ${\bf p}^O_i \in \mathbb{R}^K$.
Then the homogeneity loss can be formulated as
\begin{equation}
\label{eq:loss_homogeneity}
\setlength{\abovedisplayskip}{3pt}
\setlength{\belowdisplayskip}{3pt}
  \mathcal{L}_{h} = \left\|\frac{1}{B}\sum_{i=1}^{B}{\bf p}^O_i\right\|_2,
\end{equation}
where $B$ is the mini-batch size.
And based on the observation that an action usually occupies a small portion of a untrimmed video, we design a sparsity loss in the sparsity mechanism to relieve the interference of background segments.
And the sparsity loss is designed as
\begin{equation}
\label{eq:sparse}
\setlength{\abovedisplayskip}{3pt}
\setlength{\belowdisplayskip}{3pt}
  \mathcal{L}_{s} = \frac{1}{B}\sum_{i=1}^{B}\left\|{\bf a}_i\right\|_1,
\end{equation}
which encourages background segments to have low similarities with all the templates.

\vspace{-2mm}
\subsection{Network Training and Inference}
\vspace{-2mm}
{\bf Training}.
For training the whole network, we compose the classification loss and three auxiliary losses as
\begin{equation}
\label{eq:loss}
\setlength{\abovedisplayskip}{3pt}
\setlength{\belowdisplayskip}{3pt}
  \mathcal{L} = \mathcal{L}_{cls} + \alpha\mathcal{L}_{d} + \beta\mathcal{L}_{h} + \gamma\mathcal{L}_{s}.
\end{equation}
where $\alpha$, $\beta$ and $\gamma$ are hyper-parameters to balance the contribution of each loss function.
To summarize, we maintain a set of templates in representation of various action units and update them in the video-level classification task.
To better guide the learning of templates,
a diversity loss and a homogeneity loss are devised to keep the variety and effectiveness of each template,
and a sparsity loss is introduced to relieve background interference.

\vspace{-1mm}
{\bf Inference.}
\label{sec:AL}
After modeling action units by the AUMN, we can localize actions by examining whether a segment belongs to a kind of action unit.
In specific, we take a two-step approach to perform localization.
First, we threshold on video-level prediction scores $\hat{\bf y}_i$ and discard categories which have confidence scores below a threshold $\eta_{cls}$.
Thereafter, for each of the remaining action categories, we apply the threshold $\eta_{act}$ on the foreground attention weight to generate action proposals.
%
%}
To assign a confidence for each proposal,
we compute the class activation sequence (CAS) ${\bf C}_i \in \mathbb{R}^{{l_i}\times{C}}$ first, where
\begin{equation}
\label{eq:T-CAM}
\setlength{\abovedisplayskip}{3pt}
\setlength{\belowdisplayskip}{3pt}
	{\bf C}_i(t,;) = {\bf X}^s_i(t){\bf W}^{cls}_i(t),
\end{equation}
then ${\bf C}_i$ is passed through a softmax function along the category dimension to get class scores at each time location, denoted as ${\bf \bar C}_i$. And the confidence score $q$ in the proposal $\{(c, s, e ,q)\}$ is computed as
\begin{equation}
\label{eq:q}
\setlength{\abovedisplayskip}{3pt}
\setlength{\belowdisplayskip}{3pt}
	q = \sum_{t=s}^{e}\frac{\theta{\bf a}^{R}_i(t){\bf \bar C}^{R}_i(t,c) + (1-\theta){\bf a}^{F}_i(t){\bf \bar C}^{F}_i(t,c)}{s-e+1},
\end{equation}
where the superscripts $R$ and $F$ denote $RGB$ or $FLOW$ streams respectively, $\theta$ is a scalar denoting the relative importance between the two modalities and is set to 0.3 in this work. To remove proposals with a high overlap, the class-wise Non-Maximal Suppression (NMS) is used.

\vspace{-7pt}
\section{Experiment}
\begin{table*}[t]
\footnotesize
\centering
\caption{
Localization performance comparison with state-of-the-art methods on the THUMOS14 test set.
Note that weak\textsuperscript{+} represents methods that utilize external supervision information besides from video labels,~\emph{i.e.}, frequency of action instances.
%
%The AVG indicates the average mAP at IoU thresholds 0.1:0.1:0.5.
%
}
\label{tab:detection result on thumos14}
\vspace{5pt}
\begin{tabular}{c|c|c|cccccccc}
\hline
\multirow{2}{*}{Supervision}              & \multirow{2}{*}{Method}                                       & \multirow{2}{*}{Feature}     & \multicolumn{8}{c}{mAP@IoU} \\ \cline{4-11}
\multirow{6}{*}{Fully}     &                                                    &              & 0.1  & 0.2  & 0.3  & 0.4 & 0.5 & 0.6    & 0.7  & AVG (0.1:0.1:0.5) \\ \hline  \hline
                           & S-CNN~\cite{cvpr2016msc}, CVPR2016                 & -            & 47.7 & 43.5 & 36.3 & 28.7 & 19.0 & -    & -    & 35.0 \\
                           & R-C3D~\cite{iccv2017r-c3d}, ICCV2017               & -            & 54.5 & 51.5 & 44.8 & 35.6 & 28.9 & -    & -    & 43.1 \\
                           & SSN~\cite{iccv2017ssn}, ICCV2017                   & -            & 66.0 & 59.4 & 51.9 & 41.0 & 29.8 & -    & -    & 49.6 \\
                           & TAL-Net~\cite{cvpr2018tal-net}, CVPR2018           & -            & 59.8 & 57.1 & 53.2 & 48.5 & 42.8 & 33.8 & 20.8 & 52.3 \\
                           & GTAN~\cite{cvpr2019gaussian}, CVPR2019           & -            & 69.1 & 63.7 & 57.8 & 47.2 & 38.8 & - & - & 55.3 \\ \hline

\multirow{2}{*}{Weakly\textsuperscript{+}}
                           & STAR~\cite{aaai2019star}, AAAI2019                 & I3D          & 68.8  & 60.0 & 48.7 & 34.7 & 23.0 & -    & -    & 47.0 \\
                           & 3C-Net~\cite{iccv20193cNet}, ICCV2019              & I3D          & 59.1  & 53.5 & 44.2  & 34.1 & 26.6 & -    & 8.1  & 43.5 \\ \hline
\multirow{25}{*}{Weakly} & UntrimmedNet~\cite{cvpr2017untrimmednets}, CVPR2017    & -   & 44.4 & 37.7 & 28.2 & 21.1 & 13.7 & -    & -    & 29.0 \\
                       & Hide-and-Seek~\cite{iccv2017hide-and-seek}, ICCV2017   & -   & 36.4 & 27.8 & 19.5 & 12.7 & 6.8  & -    & -    & 20.6 \\
                       & Zhong et al.~\cite{MM2018step-by-step-erase}, MM2018   & -   & 45.8 & 39.0 & 31.1 & 22.5 & 15.9 & -    & -    & 30.9 \\
                       %& STPN~\cite{cvpr2018stpn}, CVPR2018                     & UNT & 45.3 & 38.8 & 31.1 & 23.5 & 16.2 & 9.8  & 5.1  & 31.0 \\
                       & AutoLoc~\cite{eccv2018autoloc}, ECCV2018               & UNT & -    & -    & 35.8 & 29.0 & 21.2 & 13.4 & 5.8  & -    \\
                       %& WTALC~\cite{eccv2018wtalc}, ECCV2018                   & UNT & 49.0 & 42.8 & 32.0 & 26.0 & 18.8 & -    & 6.2  & 33.7 \\
                       & Clean-Net~\cite{iccv2019weakly}, ICCV2019              & UNT & -    &  -   & 37.0 & 30.9 & 23.9 & 13.9 & 7.1  &  -   \\
                       %& CMCS~\cite{cvpr2019cmcs}, CVPR2019                     & UNT & 53.5 & 46.8 & 37.5 & 29.1 & 19.9 & 12.3 & 6.0  & 37.4 \\
                       \cline{2-11}

                       & STPN~\cite{cvpr2018stpn}, CVPR2018                     & I3D & 52.0 & 44.7 & 35.5 & 25.8 & 16.9 & 9.9  & 4.3  & 35.0 \\
                       & WTALC~\cite{eccv2018wtalc}, ECCV2018                   & I3D & 55.2 & 49.6 & 40.1 & 31.1 & 22.8 & -    & 7.6  & 39.8 \\
                       & CMCS~\cite{cvpr2019cmcs}, CVPR2019                     & I3D & 57.4 & 50.8 & 41.2 & 32.1 & 23.1 & 15.0 & 7.0  & 40.9 \\
                       & ASSG~\cite{MM2019assg}, MM2019                         & I3D & 55.6 & 49.5 & 41.1 & 31.5 & 20.9 & 13.7 & 5.9  & 39.7 \\

                       & TSM~\cite{iccv2019temporal}, ICCV2019                  & I3D & -    &  -   & 39.5 & 31.9 & 24.5 & 13.8 & 7.1  & -    \\
                       & Nguyen et al.~\cite{iccv2019backgroundmodel}, ICCV2019 & I3D & 60.4 & 56.0 & 46.6 & 37.5 & 26.8 & 19.6 & 9.0  & 45.5 \\

                       & TCAM~\cite{cvpr2020TCAM}, CVPR2020                     & I3D & -    & -    & 46.9 & 38.9 & 30.1 & 19.8 & 10.4 & -    \\
                       & DGAM~\cite{cvpr2020weakly}, CVPR2020                   & I3D & 60.0 & 54.2 & 46.8 & 38.2 & 28.8 & 19.8 & 11.4 & 45.6 \\

                       & BaS-Net~\cite{aaai2020background}, AAAI2020            & I3D & 58.2 & 52.3 & 44.6 & 36.0 & 27.0 & 18.6 & 10.4 & 43.6 \\
                       &  RPN~\cite{aaai2020relational}, AAAI2020      & I3D & 62.3 & 57.0 & 48.2 & 37.2 & 27.9 & 16.7 & 8.1  & 46.5 \\
                       & EM-MIL~\cite{eccv2020EM-MIL}, ECCV2020                 & I3D & 59.1 & 52.7 & 45.5 & 36.8 & 30.5 & \bf 22.7 & \bf 16.4  & 44.9 \\
                       &  A2CL-PT~\cite{eccv2020A2CL-PT}, ECCV2020              & I3D & 61.2 & 56.1 & 48.1 & 39.0 & 30.1 & 19.2 & 10.6  & 46.9 \\
                       & TSCN~\cite{eccv2020TSCN}, ECCV2020                     & I3D & 63.4 & 57.6 & 47.8 & 37.7 & 28.7 & 19.4 & 10.2 & 47.0 \\ \cline{2-11}

                       & AUMN                         & I3D & \bf 66.2 & \bf 61.9 & \bf 54.9 & \bf 44.4 & \bf 33.3 & 20.5 & 9.0  & \bf 52.1  \\ \hline
\end{tabular}
\vspace{-12pt}
\end{table*}

\vspace{-2mm}
\subsection{Experimental Setup}
\vspace{-2mm}
\label{sec:Experimental Setup}

{\bfseries Datasets.} The proposed AUMN is evaluated on two benchmark datasets including  THUMOS14~\cite{cviu2017thumos} and ActivityNet~\cite{cvpr2015activitynet}.
{\bfseries THUMOS14} dataset contains 200 validation videos and 213 testing videos annotated with temporal action boundaries belonging to 20 categories.
This dataset is
particularly challenging as it consists of very long videos with multiple action
instances of small duration. 
%In addition, some videos contain action instances from different categories.
%
Following previous works~\cite{cvpr2017untrimmednets,cvpr2018stpn,eccv2018wtalc,iccv2019weakly,iccv20193cNet,aaai2020background,eccv2020TSCN}, we use the 200 validation videos for training and the 213 testing videos for evaluation.
{\bfseries ActivityNet} dataset includes ActivityNet1.2 and ActivityNet1.3.
ActivityNet1.3 consists of 10024 training videos, 4926 validation videos and 5044 testing videos belonging to 200 action categories.
And ActivityNet1.2 is a subset of ActivityNet1.3, which covers 100 action categories with 4819 training, 2383 validation and 2480 testing videos.
ActivityNet only contains 1.5 instances per video on average and most videos only contain
one action category with only 36\% background averagely.
Following previous works~\cite{cvpr2017untrimmednets,cvpr2018stpn,eccv2018wtalc,iccv2019weakly,iccv20193cNet,aaai2020background,eccv2020TSCN}, we train our model on the training set and evaluate it on the validation set.
\vspace{-1mm}
{\bfseries Evaluation Metrics.} Following the standard evaluation protocol, we evaluate the TAL performance with the mean Average Precision (mAP) values under different intersection over union (IoU) thresholds.
%For ActivityNet, We use the evaluation code \footnote{https://github.com/activitynet/ActivityNet/tree/master/Evaluation} to perform the experiments.
%

%
\vspace{-1mm}
{\bfseries Implementation Details}.
We use the two-stream I3D networks~\cite{cvpr2017i3d} pre-trained on Kinetics as our feature extractor. Note that for fair comparison, we do not finetune the I3D network. 
We apply the TV-L1 algorithm to extract optical flow from RGB data. Then we divide both streams into non-overlapping 16 frames segments as the input to the I3D network, the dimension $D$ of the output feature for each segment is 1024. 
We train separate AUMNs for RGB and FLOW streams and collect the generated proposals from both networks during inference.
In AUMN, the embedding layer is composed of a temporal convolutional layer with 1024 input channels and 512 output channels.
The number of templates $K$ is $7$ if not mentioned specifically.
In Eq.~\eqref{eq:loss}, the loss function weights $\alpha$ = 0.01, $\beta$ = 0.02, and $\gamma$ is set to 0.05 and 0.03 for the RGB stream and the FLOW stream, respectively.
During inference, the threshold $\eta_{cls}$ is 0.1 and $\eta_{act}$ is the mean value of the corresponding foreground attention ${\bf a}_i$ for video $v_i$.
And we use the class-wise NMS with a threshold 0.3 to remove highly overlapped proposals.
Our model is trained using Adam optimizer~\cite{arxiv2014adam} with the learning rate $10^{-4}$ and batch size 32.

\vspace{-2mm}
\subsection{Comparison with State-of-the-art Methods}
\vspace{-2mm}
{\bf Experiments on THUMOS14}.
Table~\ref{tab:detection result on thumos14} summarizes the performance comparison between the proposed AUMN and state-of-the-art TAL methods on the THUMOS14 test set.
Weakly\textsuperscript{+} denotes methods that adopt additional supervision during training, \emph{e.g.}, the number of action instances in a video,
and AVG indicates the average mAP for IoU thresholds 0.1:0.1:0.5.
From the results, we can see that the proposed AUMN outperforms all the previous weakly supervised models and achieves a new state-of-the-art performance (33.3\% mAP at IoU0.5).
And an absolute gain of 5.1\% is achieved in terms of the average mAP when compared
to the best previous method (TSCN~\cite{eccv2020TSCN}).
It is worth noting that EM-MIL~\cite{eccv2020EM-MIL} gets a higher mAP at IoU thresholds 0.6 and 0.7 than ours.
However, we get 7\% improvement than EM-MIL at average mAP.
Besides, EM-MIL adopts a pseudo label scheme to relieve background interference while we adopt a simple sparsity prior.
We believe the performance of our approach can be promoted further when equipped with a more effective background suppression techniques.
Compared to the weakly\textsuperscript{+} methods, our method outperforms 3C-Net~\cite{iccv20193cNet} at all IoU thresholds and achieves 5.1\% improvement over STAR~\cite{aaai2019star} in average mAP.
When compared with fully supervised methods,
we note that the performance of AUMN drops faster than fully supervised methods as the IoU threshold increases.
However, we can also get a comparable result at low IoU thresholds,
e.g., AUMN outperforms TAL-Net at IoU thresholds 0.1, 0.2 and 0.3.

\vspace{-1mm}
{\bf Experiments on ActivityNet}.
On the ActivityNet dataset,
we follow the standard evaluation protocol~\cite{cvpr2015activitynet} by reporting the average mAP scores at different thresholds (0.5:0.05:0.95).
The performance comparisons on the ActivityNet1.2 and ActivityNet1.3 are shown in Table~\ref{tab:detection result on ActivityNet1.2 validation set} and Table~\ref{tab:detection result on ActivityNet1.3 validation set}, respectively.
The results are consistent with those on the THUMOS14 dataset,
and our AUMN outperforms all previous weakly supervised models in average mAP on both ActivityNet1.2 and ActivityNet1.3,
with 25.5\% and 23.5\% average mAP respectively.
It is worth noting THUMOS14 dataset and ActivityNet dataset have different characteristics.
For the THUMOS14 dataset, the most important thing is the background suppression.
While for the ActivityNet dataset, the most important thing is the localization completeness.
At high IoU thresholds,
EM-MIL has better performance on the THUMOS14 dataset while worse performance on the ActivityNet dataset.
This is because EM-MIL mainly considers background suppression while ignores the localization completeness.
Different from existing methods, our AUMN takes both background suppression and
localization completeness into consideration,
and can achieve favorable performance on both datasets.
\begin{table}[]
\centering
\footnotesize
\caption{
  Localization performance comparison with state-of-the-art methods on the ActivityNet1.2 validation set. %The AVG indicates the average mAP at IoU thresholds 0.5:0.05:0.95.
}
\label{tab:detection result on ActivityNet1.2 validation set}
\vspace{2pt}
\begin{tabular}{c|cccc}
	\hline
\multirow{2}{*}{Method}                                      &\multicolumn{4}{c}{mAP@IoU}  \\ \cline{2-5}
                                             & 0.5           & 0.75          & 0.95     & AVG \\\hline \hline
UntrimmedNet~\cite{cvpr2017untrimmednets}    & 7.4           & 3.9           & 1.2      & 3.6 \\
Zhong et al.~\cite{MM2018step-by-step-erase} & 27.3          & 14.7          & 2.9      & 15.6 \\
AutoLoc~\cite{eccv2018autoloc}               & 27.3          & 15.1          & 3.3      & 16.0 \\
WTALC~\cite{eccv2018wtalc}                   & 37.0          & 14.6          & -        & 18.0 \\
TSM~\cite{iccv2019temporal}                  & 28.3          & 17.0          & 3.5      &  -   \\
CMCS~\cite{cvpr2019cmcs}                     & 36.8          & 22.0          & 5.6      & 22.4 \\
Clean-Net~\cite{iccv2019weakly}              & 37.1          & 20.3          & 5.0      & 21.6 \\
3C-Net~\cite{iccv20193cNet}              	 & 37.2          & 23.7          & -      & 21.7 \\
Bas-Net~\cite{aaai2020background}            & 38.5          & 24.2          & 5.6      & 24.3 \\
Huang et al.~\cite{aaai2020relational}       & 37.6          & 23.9          & 5.4      & 23.3 \\
TCAM~\cite{cvpr2020TCAM}                     & 40.0          & 25.0          & 4.6      & 24.6 \\
DGAM~\cite{cvpr2020weakly}                   & 41.0          & 23.5          & 5.3      & 24.4 \\
EM-MIL~\cite{eccv2020EM-MIL}                 & 37.4          & 23.1          & 2.0      & 20.3 \\
TSCN~\cite{eccv2020TSCN}                     & 37.6          & 23.7          & \bf 5.7      & 23.6 \\

\hline
AUMN (Our's)                                 & \bf 42.0          & \bf 25.0          & 5.6      & \bf 25.5 \\
\hline
\end{tabular}
\vspace{-18pt}
\end{table}
\begin{table}[]
\centering
\footnotesize
\caption{
  Localization performance comparison with state-of-the-art methods on the ActivityNet1.3 validation set. %The AVG indicates the average mAP at IoU thresholds 0.5:0.05:0.95.
}
\label{tab:detection result on ActivityNet1.3 validation set}
\vspace{3pt}
\begin{tabular}{c|cccc}
	\hline
\multirow{2}{*}{Method}       &\multicolumn{4}{c}{mAP@IoU}  \\ \cline{2-5}
                                             & 0.5 & 0.75 & 0.95 & AVG \\ \hline \hline
STPN~\cite{cvpr2018stpn}                     & 29.3 & 16.9 & 2.6 & 16.3 \\
ASSG ~\cite{MM2019assg}                      & 32.3 & 20.1 & 4.0 & 18.8 \\
CMCS~\cite{cvpr2019cmcs}                     & 34.0 & 20.9 &\bf  5.7 & 21.2 \\
STAR~\cite{aaai2019star}                     & 31.1 & 18.8 & 4.7 & 18.2 \\
TSM~\cite{iccv2019temporal}                  & 30.3 & 19.0 & 4.5 & -    \\
Nguyen et al.~\cite{iccv2019backgroundmodel} & 36.4 & 19.2 & 2.9 & 19.5 \\
Bas-Net~\cite{aaai2020background}            & 34.5 & 22.5 & 4.9 & 22.2 \\
TSCN~\cite{eccv2020TSCN}                     & 35.3 & 21.4 & 5.3 & 21.7 \\
A2CL-PT~\cite{eccv2020A2CL-PT}               & 36.8 & 22.0 & 5.2 & 22.5 \\
                                             \hline
AUMN                                &\bf 38.3      &\bf 23.5      &5.2     &\bf 23.5     \\
                                             \hline
\end{tabular}
\vspace{-7pt}
\end{table}

\begin{table}[ht]
	\centering
	\footnotesize
	\caption{
		Ablation studies on the  THUMOS14 dataset,
		where $\mathcal{L}_{s}$, $\mathcal{L}_{d}$, $\mathcal{L}_{h}$ denote  the sparsity loss, the diversity loss and the homogeneity loss. Here, $S$ denotes the self-attention module.
	}
	\vspace{-10pt}
	\label{tab:ablation}
	\vspace{10pt}
	\setlength{\tabcolsep}{1.2mm}{
		\begin{tabular}{c|c|c|c|cccccccc}
			\hline
			\multirow{2}{*}{$\mathcal{L}_{s}$}  & \multirow{2}{*}{$\mathcal{L}_{d}$}   & \multirow{2}{*}{$\mathcal{L}_{h}$}    & \multirow{2}{*}{$S$}  & \multicolumn{8}{c}{mAP@IoU} \\ \cline{5-12}
			&     &    &      	& 0.1  & 0.2  & 0.3  & 0.4 & 0.5 & 0.6    & 0.7  & AVG \\ \hline \hline
			-	&  -    & -   & -       	& 58.5 & 53.1 & 45.1 & 34.9 & 24.6 & 14.4   & 6.8    & 43.2  \\
			\checkmark	&  -    & -   & -        & 65.1 & 59.8 & 51.5 & 40.9 & 28.8 & 16.3   & 7.3    & 49.2\\	
			\checkmark	&  \checkmark    & -   & -        & 65.8 & 61.0 & 52.6 & 42.2 & 29.5 & 17.0    & 7.6   & 50.2 \\
			\checkmark	&  -    & \checkmark   & -        & 65.5 & 60.9 & 51.7 & 41.3 & 29.4 & 17.1    & 7.7    & 49.8 \\
			\checkmark	&  \checkmark    & \checkmark  &- 		& 66.1  & 61.5 & 54.4 & 43.3 & 31.8 & 19.1 & 8.9       & 51.4\\
			\checkmark	& \checkmark    & \checkmark   & \checkmark 	& \bf 66.2 & \bf 61.9 & \bf 54.9 & \bf 44.4 & \bf 33.3 & \bf 20.5 & \bf 9.0  & \bf 52.1  \\ \hline
	\end{tabular}}
	\vspace{-15pt}
\end{table}

\vspace{-2mm}
\subsection{Ablation Studies}
\vspace{-2mm}
In this section, we conduct a series of ablation studies on the THUMOS14 dataset to evaluate the influence of each design.

\vspace{-1mm}
{\bf Influence of Each Loss Function}.
As introduced in Section~\ref{sec:aum},
we design three auxiliary losses (diversity loss $\mathcal{L}_d$, homogeneity loss $\mathcal{L}_h$ and sparsity loss $\mathcal{L}_s$) to guide the memory updating.
To explore the influence of each loss function,
we conduct experiments with different loss combinations,
and the results are shown in Table~\ref{tab:ablation}.
From the results, we have the following observations:
(1) The sparsity loss $\mathcal{L}_{s}$ can bring a significant performance improvement at all IoU thresholds.
Because there is no frame-level label as the supervision of the foreground attention,
$\mathcal{L}_{s}$ can serve as a prior to guide the action unit templates to focus on the action related segments.
(2) The diversity loss $\mathcal{L}_{d}$ is designed to encourage the action unit templates in the memory bank to be different from each other.
Without the diversity loss, we can only rely on the random initialization to achieve our goal.
From the results, we can see that the diversity loss can bring a 1.0\% performance
gain in average mAP,
which indicates that the diversity loss is necessary.
(3) The homogeneity loss $\mathcal{L}_{h}$ is designed to guarantee that each learned action unit template is useful.
Without the $\mathcal{L}_{h}$, some templates in the memory bank may be useless,
which may decrease the representative ability of the memory bank.
When equipped with this loss, we observe a 0.6\% performance gain in average mAP.
(4) These three losses can promote each other.
%
%they can work together with the classification loss to guide the learning of the memory bank.
%
For example,
the $\mathcal{L}_{d}$ can keep the difference between templates, but cannot ensure each learned template is useful.
On the other hand, although $\mathcal{L}_{d}$ can ensure no template is redundant, it may lead to learning a set of identical memory templates.
By combining them together, the mAP is increased by 3\% at IoU = 0.5,
which is much more significant than applying them independently.

\begin{figure}
	\centering
	% Requires \usepackage{graphicx}
	\includegraphics[width=\linewidth]{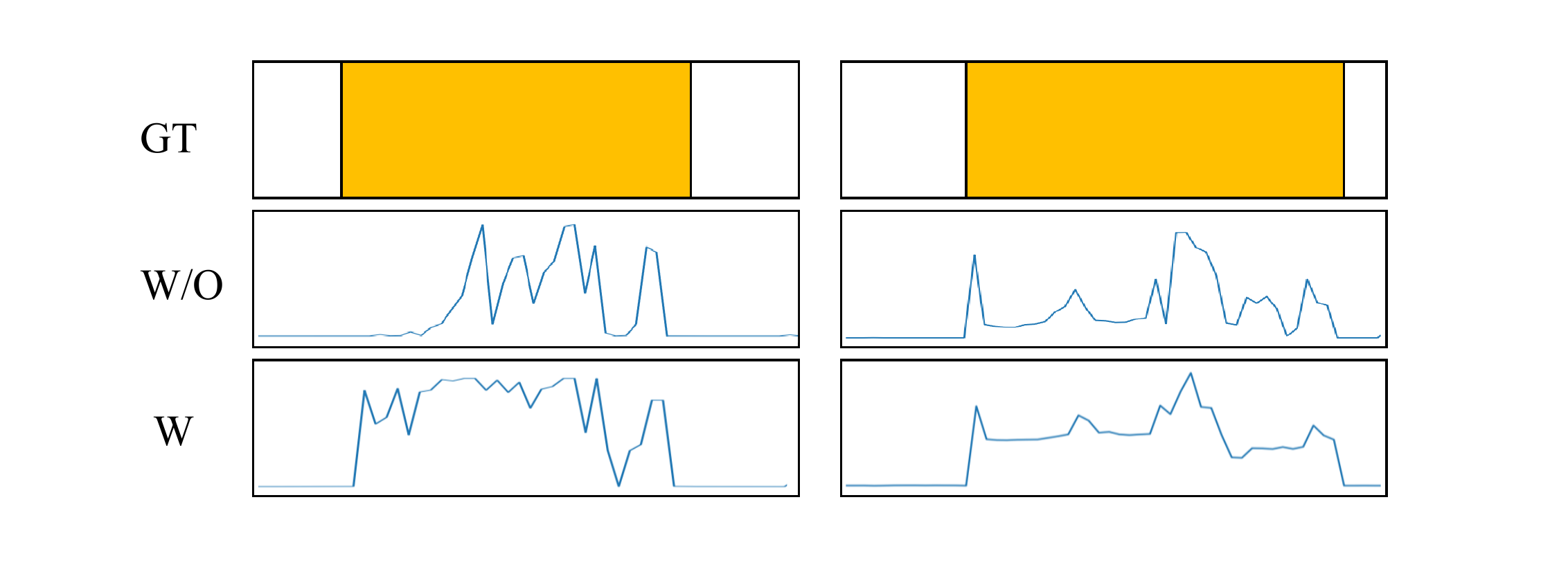}
	\caption{
		Illustration of the class activation sequence with (W) and without (W/O) the self-attention module.
	}
	\label{fig:self-attention}
\vspace{-10pt}
\end{figure}

\vspace{-1mm}
{\bf Influence of the Self-attention Module.}
As introduced in Section~\ref{sec:aum},
the self-attention module is designed to incorporate context information so as to encourage a smoother temporal classification score,
which is important for complete action localization.
From the results in Table~\ref{tab:ablation},
the self-attention module can consistently improve the performance at all IoU thresholds.
And it is worth noting that the performance gain at IoU = 0.4, 0.5, 0.6 is more significant than that at IoU = 0.1, 0.2, 0.3.
To further verify the self-attention design,
we show several visualization results in Figure~\ref{fig:self-attention}.
With the self-attention module,
some less discriminative action segments can be assigned higher confidence scores,
and it means that the self-attention module can indeed help to improve localization completeness.

\begin{figure}
	\centering
	% Requires \usepackage{graphicx}
	\includegraphics[width=0.8\linewidth]{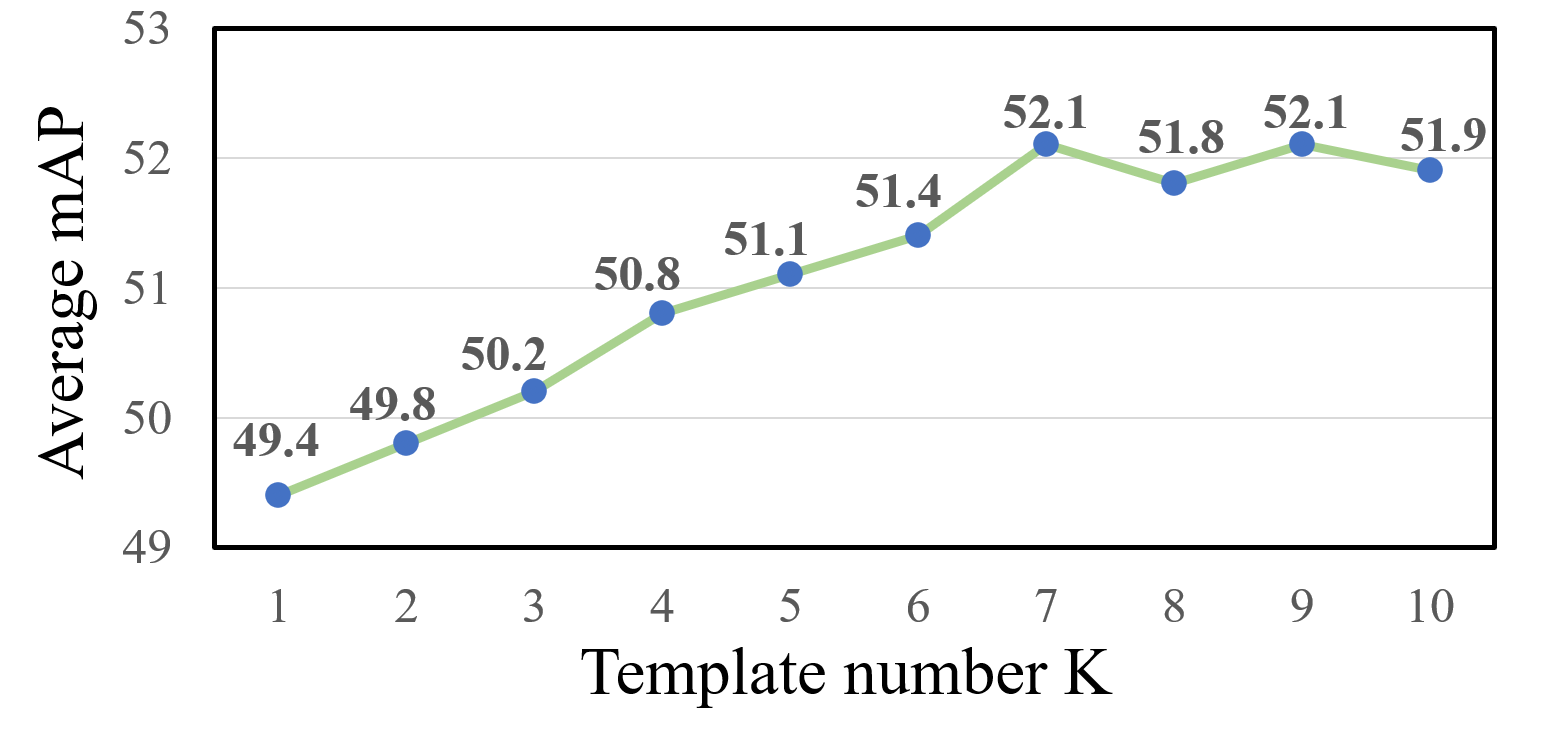}
	\caption{
		Performance comparison  of different template  numbers. The average mAP is computed at IoU thresholds 0.1:0.1:0.5. Note that the $\mathcal{L}_{d}$ and $\mathcal{L}_{h}$ are removed when K = 1.
	}
	\label{fig:ablationk}
\vspace{-15pt}
\end{figure}

\vspace{-1mm}
{\bf Influence of the Template Number.}
To explore the influence of the template number, we conduct experiments on the THUMOS14  dataset and report the average mAP at IoU 0.1:0.1:0.5 of AUMN with different template numbers.
The results are shown in Figure~\ref{fig:ablationk},
%
%{\color{red}
the average mAP can be consistently improved as the template number $K$ grows from 1 to 7,
which means 7 templates are sufficient to model all the action units on the THUMOS14
dataset.

\vspace{-2mm}
\subsection{Qualitative Results}
\begin{figure}
	\centering
	% Requires \usepackage{graphicx}
	\includegraphics[width=\linewidth]{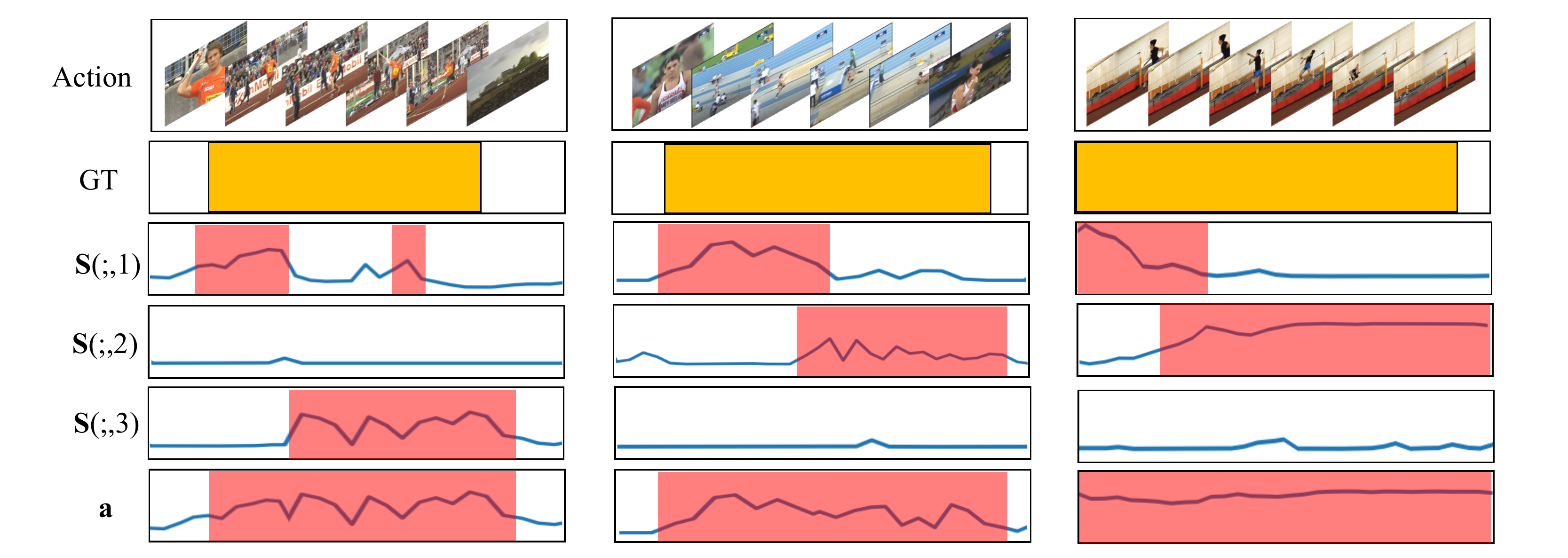}
	\caption{Qualitative results on ActivityNet1.2. The action instances from left to right are \emph{javelin-throw}, \emph{long-jump} and \emph{high-jump} respectively. The six rows in each example are input video, ground truth action instance, three different subsets of similarities scores ${\bf S}$ in Eq.~\eqref{eq:cross attention} and the foreground attention ${\bf a}$.
	}
	\label{fig:visual}
\vspace{-15pt}
\end{figure}

\vspace{-2mm}
%{\color{red}
To better understand our method, the qualitative results of our AUMN on three videos from the ActivityNet1.2 validation set are presented in Figure~\ref{fig:visual}. The action instances from left to right are \emph{javelin-throw}, \emph{long-jump} and \emph{high-jump} respectively. We visualize the similarities between video segments with different templates in third to fifth rows. And the 6th row is the foreground attention ${\bf a}$.
We find that different templates attend to model different visual patterns. For example, the first template has a high similarity to the segments which contain  the action unit running while the second is similar to jumping. The third template tends to focus on throwing, which is an important action unit in \emph{javelin-throw}. Interestingly, some segments of throwing are a little similar to the first template, because the man still keeps running while throwing the javelin.
It is worth noting that \emph{long-jump} and \emph{high-jump} both contain segments about jumping, to distinguish them from each other, the segment-wise classifiers defined in Eq~\eqref{eq:message retrieval} are desired.
In summary, by finding action units in untrimmed videos via the templates from memory and utilizing the segment-wise classifiers,
we can correctly recognize an action and obtain robust foreground attentions for complete action localization.
%}
%

\vspace{-7pt}
\section{Conclusion}
\vspace{-7pt}
In this paper, we propose an Action Unit Memory Network (AUMN) to model action units for weakly supervised temporal action localization. We design a memory bank to store the appearance and motion information of action units and their corresponding classifiers. We further introduce a cross-attention module to read segment-wise classifiers from the memory and a self-attention module for refining features by aggregating temporal context information. 
Then we can get segment-level predictions and update the memory in an adaptive way with three auxiliary mechanisms (diversity, homogeneity and sparsity). With a meaningful memory bank, we can achieve more complete localization results by finding action units in untrimmed videos.
Extensive experimental results  on two benchmarks  demonstrate  the effectiveness of the proposed AUMN.

\vspace{-7pt}
\section{Acknowledgement}
\vspace{-7pt}
This work was funded by JD AI Research,
and partially supported by the National Key Research and Development Program under Grant No. 2017YFC0820600,
Strategic Priority Research Program of Chinese Academy of Sciences (No.XDC02050500),
National Defense Basic Scientific Research Program (JCKY2020903B002),
National Nature Science Foundation of China (Grant 62022078, 62021001, 62071122),
Open Project Program of the National Laboratory of Pattern Recognition (NLPR) under Grant 202000019,
and Youth Innovation Promotion Association CAS 2018166.

{\small
\bibliographystyle{ieee_fullname}
\bibliography{egbib}
}

\end{document}